\documentclass{article}

\usepackage[nonatbib,preprint]{neurips_2022}
\usepackage{graphicx} 
\usepackage{xcolor}
\usepackage{biblatex}
\usepackage[utf8]{inputenc} 
\usepackage[T1]{fontenc}    
\usepackage{hyperref}       
\usepackage{url}            
\usepackage{booktabs}       
\usepackage{amsfonts}       
\usepackage{nicefrac}       
\usepackage{microtype}      
\usepackage{xcolor}         
\usepackage{xspace}

\newcommand{\iadet}{\texttt{IAdet}\xspace}

\title{IAdet: Simplest human-in-the-loop object detection}

\author{Franco Marchesoni-Acland\\
\And
Gabriele Facciolo\\
\And
\\
Université Paris-Saclay, ENS Paris-Saclay, CNRS, Centre Borelli, 91190, Gif-sur-Yvette, France\\
\texttt{\{marchesoniacland,gfacciol\}@gmail.com}
}

\begin{document}

\maketitle

\begin{abstract}
This work proposes a strategy for training models while annotating data 
named Intelligent Annotation (IA). IA involves three modules: (1) assisted data annotation, (2) background model training, and (3) active selection of the next datapoints. Under this framework, we open-source the \iadet tool\footnote{https://github.com/franchesoni/IAdet}, which is specific for single-class object detection.
Additionally, we devise a method for automatically evaluating such a human-in-the-loop system.
For the PASCAL VOC dataset, the \iadet tool reduces the database annotation time by $25\%$ while providing a trained model for free.
These results are obtained for a deliberately very simple \iadet design. As a consequence, \iadet is susceptible to multiple easy improvements, paving the way for powerful human-in-the-loop object detection systems. 
\end{abstract}

\section{Introduction}

In the last ten years, deep supervised learning has revolutionized many areas of science and has found widespread applications \cite{dldong2021survey}.
Although the deep learning community has used overparameterized functions for much more than vanilla supervised learning \cite{brown2020language}\cite{rematas2022urban}\cite{rombach2022highdiffusion}, supervision is still of great importance for most applications \cite{brown2020language}\cite{he2022masked}.
To upgrade supervised learning, the community has developed new architectures \cite{he2016deepresnet}\cite{dosovitskiy2020imagevit}\cite{yuan2022volo}, learning methodologies (e.g. few-shot \cite{kohler2021few}, semi-supervised \cite{van2020surveysemi}, active \cite{settles2009active}, self-supervised \cite{bardes2021vicreg}, online \cite{mai2022online}), and interactive tools \cite{gaur2018video}. Although some of these methods deal with the lack of annotated data, free tools that put the different arts together from annotation to a trained model are hard to find. Such a tool would just require an annotator and (ideally little) time to obtain a solid deep learning model. 

This paper presents the simplest example of such a tool for the specific application of single-class object detection. The tool is called \iadet, where IA stands for [Intelligent, Interactive, Incremental] Annotation.
\iadet is an example of the IA framework and it is kept as simple as possible: the main goal of this work is to present the IA scheme (this Section) and an evaluation methodology (Section~\ref{sec:evaluation}). This article also shows that the simplest implementation is already valuable for the single-class object detection problem (Section~\ref{sec:results}), and that performances could be much better (Section~\ref{sec:ablations}). 

\begin{figure}
    \centering
    \includegraphics[width=0.5\linewidth]{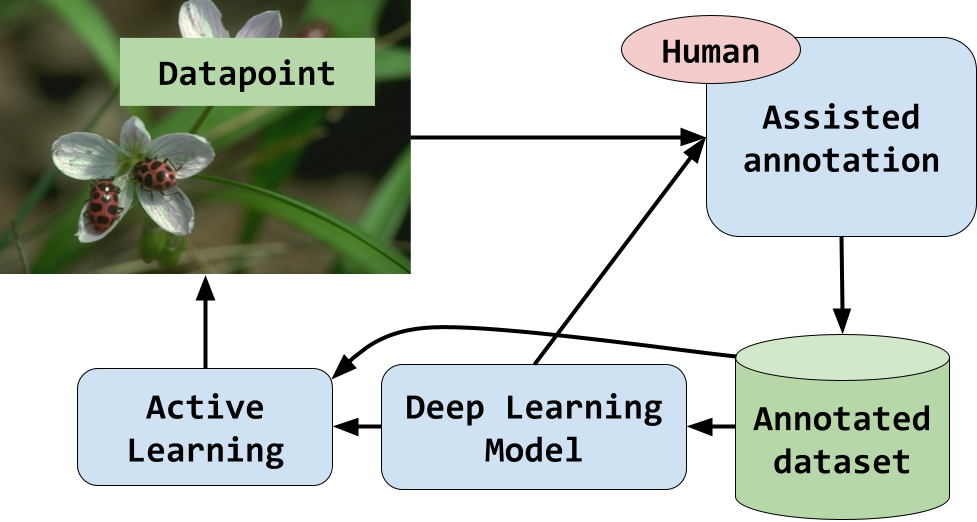}
    \caption{Conceptual IA framework.}
    \label{fig:diagram}
\end{figure}

The IA scheme comprises three components shown in Figure~\ref{fig:diagram}.
These components are not specific to object detection but are general research fields themselves. The first component is an assisted annotation tool, which allows the user to quickly make annotations. The second component is the deep learning model, which ideally achieves great performance while being very data-efficient and fast to train. The third component is an active learning method, which chooses the best next datapoint to be annotated. This loop is run until the whole dataset is annotated or the user is satisfied with the model.

Even though IA is applied in spirit, formulating such annotating-while-training systems opens new research challenges. These are the same old challenges of active learning, model training, and interactive tooling, but augmented by the interactions between them and the time and data constraints involved. Evaluation criteria are mandatory, and ideally consider the potential applications of the system: (i) reducing the annotation time and (ii) fast model training by annotating. We propose an automated method to experimentally measure these two aspects. This evaluation method is based on the simulation of the annotator.

In summary, our main contributions are
    (i) open-sourcing the \iadet tool, which allows for fast annotation and model training while annotating, 
    (ii) conceptualizing the IA human-in-the-loop learning system and a possible evaluation methodology, and
    (iii) showing that the \iadet tool can be easily and greatly improved in multiple ways.
    
The rest of the paper is organized as follows: in Section~\ref{sec:related} we describe the related work, discussing each of the many relevant research fields in the context of the related IA modules.
In the same section, we present the current industrial tooling.
Section~\ref{sec:tool} presents the \iadet tool. Its software design and the challenges there involved are presented in further detail in the Supplementary~Material. An evaluation method is needed to sort out the many questions about optimal implementation that arise. This method is exposed in Section~\ref{sec:evaluation}.
In Section~\ref{sec:results} we apply the evaluation method to our \iadet tool and present the results over the first $10$ classes of the PASCAL VOC dataset \cite{Everingham10}. Section~\ref{sec:ablations} presents ablation studies that show that the \iadet tool can be easily and significantly improved.
Finally, we conclude our work in Section~\ref{sec:conclusions} after a discussion of limitations and possible improvements in Section~\ref{sec:discussion}.

\section{Related work}\label{sec:related}

\subsection{Active learning}
Active Learning (AL) is both an IA module and a mature research field \cite{settles2009active}.
It tackles the problem of the online creation of the optimal sequence of datapoints to be annotated, as measured by a time-performance curve, where the time is measured in number of annotated datapoints. There are active learning methods that deal with deep models \cite{ren2021survey}, some modifying them, e.g. adding auxiliary heads or using ensembles \cite{beluch2018power}, and some only using their products, e.g. using the entropy of their predictions. Even though the optimal solution for a given dataset and model exists, it is unfeasible to exhaustively search for it. AL methods can be then seen as heuristics or probabilistic formulations of this problem. The usual AL baseline is random sampling.

Active learning strategies that handle neural networks usually operate in the batch setting and can be divided into diversity and uncertainty sampling. Diversity sampling looks for batches whose datapoints are representative of the data distribution \cite{sener2017active}. Uncertainty sampling uses the output distribution to find the most informative datapoints, while not necessarily enforcing diversity.
The previous state-of-the-art, BADGE \cite{badge}, combined both approaches. It has recently been surpassed by BAIT \cite{ash2021gone}, which not only has stronger theoretical foundations and performance but can also be extended to the regression setting. BAIT builds on classical maximum likelihood estimators and involves the computation of the Fisher information of the samples with respect to the parameters of the last layer.

Unfortunately, the reported performance of some active learning methods is not consistent. For instance, CORESET \cite{sener2017active} performs worse than the random sampling baseline (RSB) in the experiments reported in BAIT.
In computer vision, active learning has been mostly applied to classification tasks, with some recent works applying it to object detection \cite{yuan2021multiple}\cite{wu2022entropy}.
Recently, \citeauthor{munjal2022towards}\cite{munjal2022towards} thoroughly evaluated active learning algorithms for computer vision and showed that \textit{``the difference in performance between the AL methods and the RSB is much
smaller than reported in the literature''}, which is in line with \cite{parting}.
This comparison does not take into account the computational cost, the simplicity, and the adaptability to models and tasks of the methods, where the RSB is superior. 

\subsection{Assisted annotation}
A way of annotating more efficiently, complementary to AL, is to label more quickly. Methods that facilitate this are usually called ``interactive'', e.g. interactive image segmentation methods. The interactive annotation area studies the optimization of a loop between the human and the machine that takes place at the datapoint level, e.g. image level, in contrast to the IA loop that operates at the dataset level. For instance, in the interactive image segmentation (IIS) literature, a few user clicks guide a neural network that makes mask proposals \cite{sofiiuk2021reviving}. Such IIS tools can be used to efficiently annotate huge datasets \cite{benenson2019large}. 
Interactive segmentation tools are especially valuable because it is hard to create a detailed mask by using free-painting tools. However, for object detection, it is much easier to create and remove targets, which are bounding boxes.
Assistance in object detection is more important when numerous instances are present in the same image, as exemplified by \citeauthor{lee2022interactive}\cite{lee2022interactive}.

The ideal assisted annotation tool would propose reasonable annotations from the start and quickly converge to the desired annotation if corrections were needed. In this regard, we note that most IIS tools do not propose annotations from the start, whereas \iadet does. However, some IIS tools do provide assistance when correcting annotations \cite{chen2022focalclick}, while \iadet does not. 
More recently, general refinement blocks that can make fixed networks interactive \cite{lin2022generalizing} and language-guided methods \cite{ding2020phraseclick} were proposed, opening up even more possibilities.

\subsection{Model training}
For the IA framework, one would like a well-performing, data-efficient, and fast-to-train model. In other words, we want the model that achieves the best time-performance curve and thus enables the fastest annotation. Note that performance here refers to test performance, which we \textit{assume} to be statistically identical to the performance over unlabeled samples. In what follows we will restrict our attention to object detection models, but their literature is representative of other areas as well.

The best performing models are huge, thus they are not fast to train and are not necessarily data efficient. For instance, the current state-of-the-art in COCO detection \cite{wei2022contrastive} is a self-supervised transformer with 3 billion parameters. On the other hand, the most data-efficient models are presumably few-shot object detection models. Few-shot object detection models are usually classified into meta-learned models and finetuned models. Note that both are usually finetuned, and both achieve comparable performance \cite{kohler2021few}. The meta-learned models are trained with episodes that simulate the few-shot setting by using a limited number of arbitrary query vectors that have to be recognized in the training images.
In the IA framework, the amount of annotated data increases, yet meta-learned few-shot models do not scale to more annotations as naturally as the simpler finetuned models \cite{wang2020frustratingly}.
More recently, state-of-the-art few-shot detection performance has been established by leveraging self-supervised representations \cite{detreg}\cite{huang2022survey}, which suggests that traditionally fine-tuning self-supervised models can be better than using few-shot specific ones. One important lesson from the few-shot learning literature is that full finetuning is usually better than retraining the head only \cite{kohler2021few}, although simply retraining the head can bring solid performance too \cite{wang2020frustratingly}. Even if few-shot learning methods have been greatly developed, simple baselines are still competitive \cite{wang2020frustratingly}\cite{pushing}. 

The semi-supervised learning literature explores learning from little annotated data and much more unlabeled data. Self-supervised learning has enabled the extraction of supervised-level representations from unlabeled data only. These representations make further learning (even self-supervised \cite{reed2022self}) easier. 
These methods usually involve pretraining, but we prefer to save time by starting from already pretrained backbones. How to best do this is studied by transfer learning. A comparison between transfer and self-supervised learning is given in \cite{yang2020transfer}. \citeauthor{mensink2021factors}\cite{mensink2021factors} have shown that the most important factor for transfer learning is the source image domain, which should ideally include the target image domain. Moreover, there are metrics one could compute to predict which models would better transfer to a target dataset \cite{agostinelli2022transferability}. If the source and target domain are different, one should look into the domain adaptation literature \cite{wang2018deep}. In this work, we make transfer learning suitable by using MS-COCO \cite{lin2014microsoft} weights to tackle PASCAL VOC: the classes of the former dataset include those of the latter.

\subsection{Industrial tools}
There are between 15 and 20 companies offering data labeling tools for computer vision. There are also many open-source tools for annotation, e.g. \cite{russell2008labelme}\cite{dutta2019via} or \href{https://pixano.cea.fr/about/}{Pixano}. To the best of our knowledge, there are no generic free tools that involve assistance and allow training while annotating. From the tools behind a paywall, our experience suggests that the assistance, that only some of them have, is not as developed as marketed. The closest to \iadet is \cite{aide}, but it was presented as specific to ecology and does not tackle the questions of what a correct evaluation procedure is, or what happens when such a loop is constructed. The assisted annotation area is attracting more attention every day, however there are no good open tools yet.

\section{\iadet tool}\label{sec:tool}

\begin{figure}
    \centering
    \includegraphics[width=0.5\linewidth]{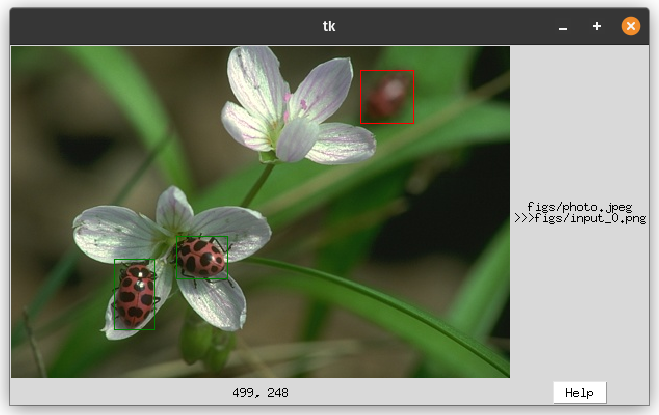}
    \caption{Graphical User Interface (GUI) of \iadet. Predictions are shown in green and an extra annotation in red. The visible components are the picture and the bounding boxes, a help button, the list of images in the directory, and the mouse coordinates.}
    \label{fig:gui}
\end{figure}
The \iadet tool is defined by its modules and their interaction. We aimed for the simplest possible tool and we selected it without any tuning. This tool can (and will) be enhanced by using better components. For the assisted annotation module, we developed a Graphical User Interface that uses the model to suggest bounding boxes before starting the annotation (see Figure~\ref{fig:gui}). For the deep learning model, we use SSD \cite{liu2016ssd}, one of the simplest and worst performing object detection architectures available in \verb|mmdet| \cite{mmdetection}, a well-maintained computer vision library. The chosen active learning algorithm is the random sampling baseline, i.e. we avoid using active learning.



The interaction between the components is done via files: the Graphical User Interface (GUI) implements assisted annotation by using the latest model weights to predict the bounding boxes for the image to be labeled next.
The annotations are saved in a file with \verb|mmdet| standardized format that contains the bounding boxes as vectors of length $4$ and the path to each corresponding image. 
Simultaneously, the background model trains continually by repeating the following operations: (i) load the annotation file and (ii) train one epoch over it. All hyperparameters are the default of the chosen library (SSD300 + PASCALVOC).


Our user experience (UX)-aware decisions involved avoiding the overlay of annotations and predictions, preferring click-click over click-drag bounding boxes, removing a bounding box with one click, removing all bounding boxes and navigating the dataset directory with the keyboard, automatically saving when leaving an image, between others. We provide a more detailed description and discussion of the design and the limitations in the supplementary material.

\section{Evaluation}\label{sec:evaluation}
A comprehensive evaluation should include the interaction between the modules, and since they are complex and evolve through time, the full annotation process needs to be simulated.
For our \iadet tool the automatic evaluation will be simple: a robot annotator will  make the  annotations  for new images drawing them from the ground truth at a rate of $R \;\texttt{interactions}/s$.
Computing the number of interactions per image is straightforward. 1 key-press is needed to move to the next image and also to erase all predicted bounding boxes. 2 clicks are needed to create one new bounding box, and 1 click is needed to remove one bounding box. Assuming perfect annotation with the least number of interactions, and given the number of false-positives (FP), false-negatives (FN), and true-positives (TP), the time to annotate the $i$th image is $I_i / R$, where 
$I_i = \left(1 + \min{\left(1+(TP+FN)\times2, FP+FN\times2 \right)}\right)$
and the $\min$ considers the best strategy between ignoring all predictions or correcting them. Note that choosing to ignore all predictions involves an extra interaction, thus it is strictly worse than not using the tool at all. If we \textit{assume} this model of annotation time as correct, an annotating robot with speed $R$ can be used to simulate a full dataset annotation in real-time.

For a given dataset and IA system, one can think of the performance as dependent on a few values: the speed of annotation $R$, the training speed $v$ and batch size $b$, and the elapsed time $t$. In other words, for each $(R,v,b)$, we can obtain a time-performance curve. This curve is defined for $t \in [0, t_A]$, where $t_A$ is the time taken to annotate the whole dataset.
Another important annotation time is $t_N = I / R$, where $I$ is the total number of needed unassisted interactions, or $2\times$ the number of bounding boxes in all ground truths. Because these times are highly dependent on the speed of annotation $R$, it is better to look at their ratio, $t_A / t_N$, which roughly measures the incremental error incurred by the model. In the ratio, $R$ seems to disappear, but it is not the case: $t_A$ continues to depend on $R$ because $R$ roughly determines how many new images the model will see at each epoch, which in turn changes its performance. We also note that $t_N$ is not necessarily an upper bound of $t_A$, as bad predictions cause one more interaction per image. 

The final model performance or negative loss $-L(t_A)$ could be compared with the one achieved by the same model but trained with the full annotated dataset instead of incrementally, which would be the ideal case. This supervised performance is $-L_{sup}$ and it is not necessarily an upper bound of $-L(t_A)$, as some active learning methods could outperform their fully supervised reference.
Naturally, this performance is evaluated on the test split of the given dataset. Analogously, we can take the performance ratio $L_A / L_{sup}$ as a key metric. This metric measures how well the background model performs at the end of the annotation relative to the performance of a model trained \textit{after} the dataset was fully annotated. Of course, one could always use the \iadet tool to annotate before training a new model, but it is interesting to know if the background model has any use after annotation.

In summary, simulating the annotation at $R$ interactions per second, we can measure the change in required annotation time $t_A/t_N$ and the relative final model performance $L_A / L_{sup}$. This evaluation protocol is also useful to evaluate the merits of individual modules, as some of their most important potential applications live inside the IA framework.




\section{Results}\label{sec:results}

Our experiments deal with the simplistic single-class object detection problem. We further \textit{assume} that all images contain the class of interest. We choose the first half of the classes in PASCAL VOC 2007 and 2012 \cite{Everingham10} for the experiment regarding the annotation time and the first three classes to analyze background-model performances. For the ablation studies, we use the least common class, \textit{sheep}, for which we have 420 training or validation examples and 97 test examples (from the PASCAL 2007 test split). Our objective is twofold: first, compare the annotation time while using assistance against the default annotation time, and second, compare the performance of the assistant model against the performance of a supervised model that had access to the fully annotated dataset. All experiments were run sequentially on a single GPU NVIDIA TITAN V.

\begin{table}
  \caption{Annotation performances per class for \iadet and $R=1\; \texttt{interactions}/s$.}
  \label{tab:time}
  \centering
  \begin{tabular}{llllllc}
    \toprule
    Index & Class & \# Images & $t_A (s)$ & $t_N (s)$ & $t_A / t_N$ & \% Improvement \\
    \midrule
0  & aeroplane   & 908    &  2196 & 3247   & 0.676 & 32.3 \\ 
1  & bicycle     & 795   &   2182 &  2920  & 0.747 & 25.2 \\
2  & bird        & 1095   &  3138 &  4302  & 0.729 & 27.0 \\
3  & boat        & 689   &   2562 &  2966  & 0.864 & 13.5 \\
4  & bottle      & 950   &   3886 &  4475  & 0.868 & 13.1 \\
5  & bus         & 607   &   1593 &  2248  & 0.709 & 29.0 \\
6  & car         & 1874   &  5618 &  8405  & 0.668 & 33.1 \\
7  & cat         & 1417   &  2770 &  4600  & 0.602 & 39.7 \\
8  & chair       & 1564   &  6219 &  7865  & 0.790 & 20.9 \\
9  & cow         & 444   &   1667 &  2129  & 0.783 & 21.6 \\
16 & sheep       & 421   &   2007 & 2558  &  0.784 & 21.5 \\
    \midrule
  & Mean    & & & & 0.747 & \textbf{25.2}\\
    \bottomrule
  \end{tabular}
\end{table}

\begin{figure}
    \centering
    \includegraphics[width=0.5\linewidth]{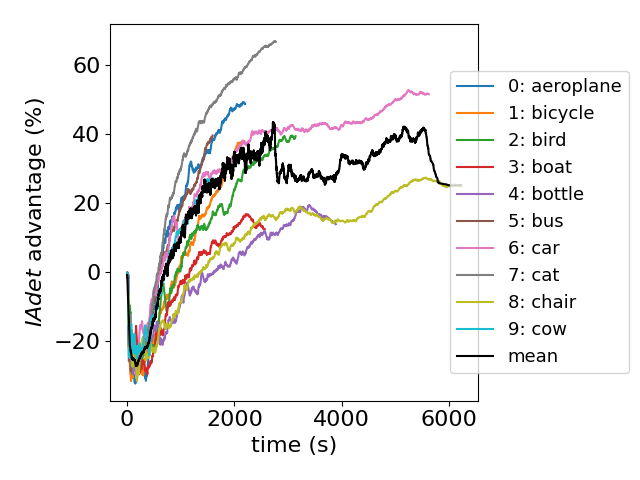}
    \caption{Advantage $k_A(t)/k_N(t)$ vs. time, where $k_A(t)$ ($k_N(t)$) is the number of annotations made before time $t$ with (without) assistance. The mean is a box filter of all curves.}
    \label{fig:ratios}
\end{figure}




Table~\ref{tab:time} presents the results for the first half of the classes of the dataset. The table shows that: (i) the unassisted annotation time is about one hour per class and (ii) when using the tool, the average improvement is $25\%$. Figure~\ref{fig:ratios} shows the increase in the advantage 
provided by the tool as time goes by. We noticed a dip caused by a first stage in which the predictions of the model are not yet useful and have to be discarded. This is likely the greatest drawback of the present tool, as it takes more than $100$ annotated images for the assistance to start being useful.



\begin{figure}
    \centering
    \includegraphics[width=0.5\linewidth]{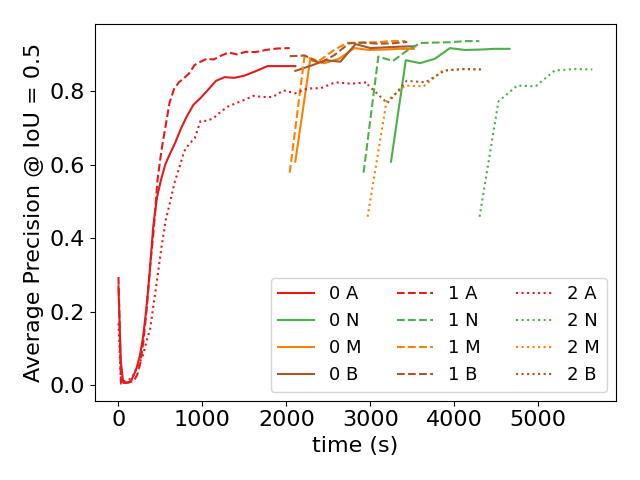}
    \caption{Test-set performance
    for the background model training variations (A, N, M, B), described in the text. The experiment was conducted for the classes 0, 1, and 2 of PASCAL VOC.
    }
    \label{fig:apvst}
\end{figure}

Closely related to the annotation speedup is the model performance through time. Figure~\ref{fig:apvst} compares the Average Precision (AP) for four cases. The first is the model trained with annotations that are incrementally added to the dataset by the (simulated) annotator (A). The second and third are models trained with all annotations, which could happen after unassisted annotation (N) or after assisted annotation (M). As more information is available in these cases, we expect the (N) or (M) models (which are identical up to a time shift) to reach higher performance, i.e. to set an upper bound for the first curve (A). The background model improves performance through time but can not reach the one of the supervised. The other option we consider, (B), is to train the model after assisted annotations with initial weights coming from (A).
From Figure~\ref{fig:apvst} one can observe that there is not much difference between bootstrapping (B) or training from scratch (M) after the first few epochs.

\begin{table}
  \caption{Final performance of models trained at different times measured by Average Precision (AP) at 0.5 IoU. (A) during annotation, (N) after unassisted annotation, and (B) after assisted annotation initialized with final weights of (A). Experiment done with the first three classes of PASCAL VOC.}
  \label{tab:performance}
  \centering
  \begin{tabular}{l|lllc}
    \toprule
    Index & A  & N  & B  & $A/N$ (\%)\\
    \midrule
0  & 0.868 & 0.915 & 0.922 & 94.8 \\
1  & 0.917 & 0.936 & 0.934 & 97.9 \\
2  & 0.824 & 0.859 & 0.859 & 95.9 \\
    \midrule
    Mean & 0.870 & 0.903 & 0.905 & 96.2\\
    \bottomrule
  \end{tabular}
\end{table}

Table~\ref{tab:performance} shows the final IoUs for the models (A, N, B) for the first three classes of PASCAL VOC. The model we get for free after the assisted annotation of a full dataset achieves a performance only $5\%$ below that of the model trained after annotation. 

\section{Ablations}\label{sec:ablations}
\begin{table}
  \caption{\iadet performance for different annotation rates $R$ for the \textit{sheep} class (16).}
  \label{tab:speed}
  \centering
  \begin{tabular}{lllllc}
    \toprule
    $R$ & description & $t_A (s)$ & $t_N (s)$ & $t_A / t_N$ & advantage (\%)\\
    \midrule
1  & baseline              &  2007 & 2558& 0.784 & 21.5 \\ 
0.2  & ($\times 5$) slower &   8059 &  12790  & 0.63 & 36.9 \\
5 & ($\times 5$) faster    &  572 & 511 & 1.11 & -11.8 \\
    \bottomrule
  \end{tabular}
\end{table}

Table~\ref{tab:speed} compares different annotation speeds. What we find is that if the annotator is slower the \iadet tool adds more value: the model learns faster relative to how much has been annotated. In the other direction, if the annotator is very fast, the \iadet tool is too slow to learn and do anything useful.

\begin{figure}
    \centering
    \includegraphics[width=0.5\linewidth]{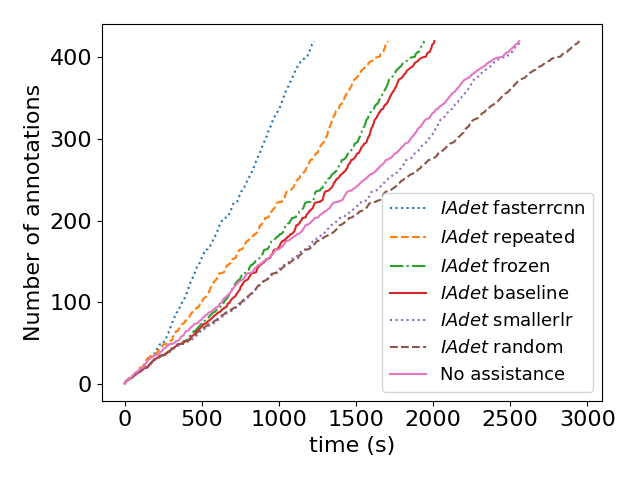}
    \caption{Ablation studies trying Faster-RCNN (fasterrcnn), repeating and augmenting the training datapoints (repeated), freezing the backbone (frozen), a $10 \times$ smaller learning rate (smallerlr), and a random initialization (random). Experiment conducted over the class \textit{sheep} of the PASCAL VOC dataset.}
    \label{fig:ablations}
\end{figure}

Figure~\ref{fig:ablations} and Table~\ref{tab:ablations} compare different changes on the training procedure. The first change, \textit{random}, which deteriorates the performance the most, is to initialize the weights of the model not from COCO, but randomly. The second change evaluated was to make the initial learning rate $10 \times$ smaller. Both these changes make the model learn less quickly to the point of making assistance counterproductive.
The third variation evaluated, \textit{frozen}, involves freezing the backbone weights coming from the COCO-trained model, but not the neck and head weights. This allows for a modest improvement, although it does not provide the fast and robust performance one could expect. The fourth change is a hack: instead of using each image once in an epoch, we use it $10$ times with various augmentations (e.g. random crop, flip, and photo-metric distortion), thereby reducing the time needed to load the dataset at each epoch and increasing data diversity. This change makes the model useful much sooner. Lastly, we change the model from SSD \cite{liu2016ssd} to a modern Faster R-CNN \cite{ren2015faster} with a Feature Pyramid Network \cite{lin2017feature} ResNet 50 \cite{he2016deepresnet} as the backbone. The architectural change is the most impactful, reducing annotation time by more than $50\%$. Note that we do not combine improvements, as this section solely aims to show that the base \iadet tool can be easily improved.


\begin{table}
  \caption{Annotation performances for variations of \iadet with $R=1$.}
  \label{tab:ablations}
  \centering
  \begin{tabular}{llllc}
    \toprule
    Ablation &  $t_A (s)$ & $t_N (s)$ & $t_A / t_N$ & \% Improvement \\
    \midrule
random          &  2948 &  2558 & 1.15 & -15.2 \\
smallerlr       &  2565 &  2558 & 1.003 & -0.3 \\
baseline        &  2007 &  2558 & 0.784 & 21.5 \\ 
frozen          &  1941 &  2558 & 0.758 & 24.1 \\
repeated        &  1705 &  2558 & 0.666 & 33.3 \\
fasterrcnn      &  1227 &  2558 & 0.479 & 52.0 \\
    \bottomrule
  \end{tabular}
\end{table}

\section{Limitations}\label{sec:discussion}

The limitations of \iadet relate to all the obvious possible improvements: 
using a better background model, e.g. Faster-RCNN \cite{ren2015faster} (as shown) or DETR \cite{carion2020end}, including an active learning algorithm, adding intra-image assistance. Other than these, the main limitations are the domain shift, which makes transfer learning and the current tool not as useful for real applications, and the long time it takes the model to produce relevant proposals. The greatest weakness of the evaluation procedure is that we did not consider the labeling noise introduced by the model, whose predictions were defined as correct with the standard $0.5$ IoU threshold. Future work could include few-shot learning methods such as \cite{wang2020frustratingly}.

\section{Conclusions}\label{sec:conclusions}
The generic IA framework for both fast-annotation and training-while-annotating was introduced in this paper, which ideally combines assistance, background model training, and active learning. This IA framework was exemplified by the \iadet tool, an extremely simple tool to make annotations for object detection. Despite its simplicity, \iadet is effective: it can reduce the annotation time by $25\%$ and provides a competitive trained model for free.
Any human-in-the-loop system similar to \iadet could be a very interesting application where many related fields of machine learning could converge to. We have shown that there is value in such systems (even when they are very simple) and that there is ample room for improvement.

While we were in the course of correcting this paper, we discovered that a previous reference \cite{icpr} had addressed a similar problematic and reached several of the conclusions presented herewith. In particular, they propose a similar human-in-the-loop system including a Faster-RCNN architecture and evaluate the annotation efficiency over datasets including PASCAL VOC, which is definitely similar to our developments. Their conclusions are also comparable to ours, in particular they reach a workload reduction between 30\% and 60\% which can be compared to our 52\% result in Table~\ref{tab:ablations}. The main differences are that IAdet i) presents no waiting time to the user, ii) was evaluated in real time, iii) assumes different number of clicks for FPs and FNs, iv) does not operate in a per-batch setting, and v) is open sourced.



\section{Potential Negative Societal Impacts}
The authors do not recognize any potentially negative societal impact of this paper other than potentially lowering the barrier of entry to object detection model training.

\begin{ack}
Work partly financed by Office of Naval research grant N00014-20-S-B001, MENRT, and by a grant from ANRT. 
Centre Borelli is also a member of Université Paris Cité, SSA and INSERM.


\end{ack}


\printbibliography

@Article{Everingham10,
  author={Everingham, Mark and Eslami, SM and Van Gool, Luc and Williams, Christopher KI and Winn, John and Zisserman, Andrew},
   title = "The Pascal Visual Object Classes (VOC) Challenge",
   journal = "International Journal of Computer Vision",
   volume = "88",
   year = "2010",
   number = "2",
   month = jun,
   pages = "303--338",
}

@techreport{settles2009active,
  title={Active learning literature survey},
  author={Settles, Burr},
  year={2009},
  institution={University of Wisconsin-Madison Department of Computer Sciences}
}

@article{ren2021survey,
  title={A survey of deep active learning},
  author={Ren, Pengzhen and Xiao, Yun and Chang, Xiaojun and Huang, Po-Yao and Li, Zhihui and Gupta, Brij B and Chen, Xiaojiang and Wang, Xin},
  journal={ACM computing surveys (CSUR)},
  volume={54},
  number={9},
  pages={1--40},
  year={2021},
  publisher={ACM New York, NY}
}

@inproceedings{beluch2018power,
  title={The power of ensembles for active learning in image classification},
  author={Beluch, William H and Genewein, Tim and N{\"u}rnberger, Andreas and K{\"o}hler, Jan M},
  booktitle={Proceedings of the IEEE conference on computer vision and pattern recognition},
  pages={9368--9377},
  year={2018}
}

@article{sener2017active,
  title={Active learning for convolutional neural networks: A core-set approach},
  author={Sener, Ozan and Savarese, Silvio},
  journal={arXiv preprint arXiv:1708.00489},
  year={2017}
}

@article{badge,
  author    = {Jordan T. Ash and
               Chicheng Zhang and
               Akshay Krishnamurthy and
               John Langford and
               Alekh Agarwal},
  title     = {Deep Batch Active Learning by Diverse, Uncertain Gradient Lower Bounds},
  journal   = {CoRR},
  volume    = {abs/1906.03671},
  year      = {2019},
  eprinttype = {arXiv},
  eprint    = {1906.03671},
  bibsource = {dblp computer science bibliography, https://dblp.org}
}

@article{ash2021gone,
  title={Gone fishing: Neural active learning with fisher embeddings},
  author={Ash, Jordan and Goel, Surbhi and Krishnamurthy, Akshay and Kakade, Sham},
  journal={Advances in Neural Information Processing Systems},
  volume={34},
  pages={8927--8939},
  year={2021}
}

@inproceedings{yuan2021multiple,
  title={Multiple instance active learning for object detection},
  author={Yuan, Tianning and Wan, Fang and Fu, Mengying and Liu, Jianzhuang and Xu, Songcen and Ji, Xiangyang and Ye, Qixiang},
  booktitle={Proceedings of the IEEE/CVF Conference on Computer Vision and Pattern Recognition},
  pages={5330--5339},
  year={2021}
}

@inproceedings{wu2022entropy,
  title={Entropy-based Active Learning for Object Detection with Progressive Diversity Constraint},
  author={Wu, Jiaxi and Chen, Jiaxin and Huang, Di},
  booktitle={Proceedings of the IEEE/CVF Conference on Computer Vision and Pattern Recognition},
  pages={9397--9406},
  year={2022}
}

@inproceedings{munjal2022towards,
  title={Towards robust and reproducible active learning using neural networks},
  author={Munjal, Prateek and Hayat, Nasir and Hayat, Munawar and Sourati, Jamshid and Khan, Shadab},
  booktitle={Proceedings of the IEEE/CVF Conference on Computer Vision and Pattern Recognition},
  pages={223--232},
  year={2022}
}

@inproceedings{sofiiuk2021reviving,
  title={Reviving iterative training with mask guidance for interactive segmentation},
  author={Sofiiuk, Konstantin and Petrov, Ilya A and Konushin, Anton},
  booktitle={2022 IEEE International Conference on Image Processing (ICIP)},
  pages={3141--3145},
  year={2022},
  organization={IEEE}
}

@inproceedings{benenson2019large,
  title={Large-scale interactive object segmentation with human annotators},
  author={Benenson, Rodrigo and Popov, Stefan and Ferrari, Vittorio},
  booktitle={Proceedings of the IEEE/CVF Conference on Computer Vision and Pattern Recognition},
  pages={11700--11709},
  year={2019}
}

@inproceedings{lee2022interactive,
  title={Interactive Multi-Class Tiny-Object Detection},
  author={Lee, Chunggi and Park, Seonwook and Song, Heon and Ryu, Jeongun and Kim, Sanghoon and Kim, Haejoon and Pereira, S{\'e}rgio and Yoo, Donggeun},
  booktitle={Proceedings of the IEEE/CVF Conference on Computer Vision and Pattern Recognition},
  pages={14136--14145},
  year={2022}
}

@inproceedings{chen2022focalclick,
  title={FocalClick: Towards Practical Interactive Image Segmentation},
  author={Chen, Xi and Zhao, Zhiyan and Zhang, Yilei and Duan, Manni and Qi, Donglian and Zhao, Hengshuang},
  booktitle={Proceedings of the IEEE/CVF Conference on Computer Vision and Pattern Recognition},
  pages={1300--1309},
  year={2022}
}

@inproceedings{ding2020phraseclick,
  title={Phraseclick: toward achieving flexible interactive segmentation by phrase and click},
  author={Ding, Henghui and Cohen, Scott and Price, Brian and Jiang, Xudong},
  booktitle={European Conference on Computer Vision},
  pages={417--435},
  year={2020},
  organization={Springer}
}

@inproceedings{lin2022generalizing,
  title={Generalizing Interactive Backpropagating Refinement for Dense Prediction Networks},
  author={Lin, Fanqing and Price, Brian and Martinez, Tony},
  booktitle={Proceedings of the IEEE/CVF Conference on Computer Vision and Pattern Recognition},
  pages={773--782},
  year={2022}
}

@article{wei2022contrastive,
  title={Contrastive Learning Rivals Masked Image Modeling in Fine-tuning via Feature Distillation},
  author={Wei, Yixuan and Hu, Han and Xie, Zhenda and Zhang, Zheng and Cao, Yue and Bao, Jianmin and Chen, Dong and Guo, Baining},
  journal={arXiv preprint arXiv:2205.14141},
  year={2022}
}

@article{kohler2021few,
  title={Few-Shot Object Detection: A Survey},
  author={K{\"o}hler, Mona and Eisenbach, Markus and Gross, Horst-Michael},
  journal={arXiv preprint arXiv:2112.11699},
  year={2021}
}

@InProceedings{detreg,
    author    = {Bar, Amir and Wang, Xin and Kantorov, Vadim and Reed, Colorado J. and Herzig, Roei and Chechik, Gal and Rohrbach, Anna and Darrell, Trevor and Globerson, Amir},
    title     = {DETReg: Unsupervised Pretraining With Region Priors for Object Detection},
    booktitle = {Proceedings of the IEEE/CVF Conference on Computer Vision and Pattern Recognition (CVPR)},
    month     = {6},
    year      = {2022},
    pages     = {14605-14615}
}

@article{huang2022survey,
  title={A survey of self-supervised and few-shot object detection},
  author={Huang, Gabriel and Laradji, Issam and V{\'a}zquez, David and Lacoste-Julien, Simon and Rodriguez, Pau},
  journal={IEEE Transactions on Pattern Analysis and Machine Intelligence},
  year={2022},
  publisher={IEEE}
}

@article{mensink2021factors,
  title={Factors of influence for transfer learning across diverse appearance domains and task types},
  author={Mensink, Thomas and Uijlings, Jasper and Kuznetsova, Alina and Gygli, Michael and Ferrari, Vittorio},
  journal={arXiv preprint arXiv:2103.13318},
  year={2021}
}

@inproceedings{agostinelli2022transferability,
  title={Transferability Metrics for Selecting Source Model Ensembles},
  author={Agostinelli, Andrea and Uijlings, Jasper and Mensink, Thomas and Ferrari, Vittorio},
  booktitle={Proceedings of the IEEE/CVF Conference on Computer Vision and Pattern Recognition},
  pages={7936--7946},
  year={2022}
}

@article{wang2018deep,
  title={Deep visual domain adaptation: A survey},
  author={Wang, Mei and Deng, Weihong},
  journal={Neurocomputing},
  volume={312},
  pages={135--153},
  year={2018},
  publisher={Elsevier}
}

@inproceedings{lin2014microsoft,
  title={Microsoft coco: Common objects in context},
  author={Lin, Tsung-Yi and Maire, Michael and Belongie, Serge and Hays, James and Perona, Pietro and Ramanan, Deva and Doll{\'a}r, Piotr and Zitnick, C Lawrence},
  booktitle={European conference on computer vision},
  pages={740--755},
  year={2014},
  organization={Springer}
}

@article{aide,
author = {Kellenberger, Benjamin and Tuia, Devis and Morris, Dan},
title = {AIDE: Accelerating image-based ecological surveys with interactive machine learning},
journal = {Methods in Ecology and Evolution},
volume = {11},
number = {12},
pages = {1716-1727},
doi = {10.1111/2041-210X.13489},
year = {2020}
}

@article{parting,
  doi = {10.48550/ARXIV.1912.05361},
  
  author = {Mittal, Sudhanshu and Tatarchenko, Maxim and Çiçek, Özgün and Brox, Thomas},

  title = {Parting with Illusions about Deep Active Learning},
  
  journal={arXiv preprint arXiv:1912.05361},
  
  year = {2019}
  
}

@article{wang2020frustratingly,
  title={Frustratingly simple few-shot object detection},
  author={Wang, Xin and Huang, Thomas E and Darrell, Trevor and Gonzalez, Joseph E and Yu, Fisher},
  journal={arXiv preprint arXiv:2003.06957},
  year={2020}
}

@article{bardes2021vicreg,
  title={Vicreg: Variance-invariance-covariance regularization for self-supervised learning},
  author={Bardes, Adrien and Ponce, Jean and LeCun, Yann},
  journal={arXiv preprint arXiv:2105.04906},
  year={2021}
}

@article{van2020surveysemi,
  title={A survey on semi-supervised learning},
  author={Van Engelen, Jesper E and Hoos, Holger H},
  journal={Machine Learning},
  volume={109},
  number={2},
  pages={373--440},
  year={2020},
  publisher={Springer}
}

@inproceedings{reed2022self,
  title={Self-supervised pretraining improves self-supervised pretraining},
  author={Reed, Colorado J and Yue, Xiangyu and Nrusimha, Ani and Ebrahimi, Sayna and Vijaykumar, Vivek and Mao, Richard and Li, Bo and Zhang, Shanghang and Guillory, Devin and Metzger, Sean and others},
  booktitle={Proceedings of the IEEE/CVF Winter Conference on Applications of Computer Vision},
  pages={2584--2594},
  year={2022}
}

@article{yang2020transfer,
  title={Transfer learning or self-supervised learning? A tale of two pretraining paradigms},
  author={Yang, Xingyi and He, Xuehai and Liang, Yuxiao and Yang, Yue and Zhang, Shanghang and Xie, Pengtao},
  journal={arXiv preprint arXiv:2007.04234},
  year={2020}
}

@inproceedings{pushing,
  title={Pushing the Limits of Simple Pipelines for Few-Shot Learning: External Data and Fine-Tuning Make a Difference},
  author={Hu, Shell Xu and Li, Da and St{\"u}hmer, Jan and Kim, Minyoung and Hospedales, Timothy M},
  booktitle={Proceedings of the IEEE/CVF Conference on Computer Vision and Pattern Recognition},
  pages={9068--9077},
  year={2022}
}

@article{russell2008labelme,
  title={LabelMe: a database and web-based tool for image annotation},
  author={Russell, Bryan C and Torralba, Antonio and Murphy, Kevin P and Freeman, William T},
  journal={International journal of computer vision},
  volume={77},
  number={1},
  pages={157--173},
  year={2008},
  publisher={Springer}
}

@inproceedings{dutta2019via,
  author = {Dutta, Abhishek and Zisserman, Andrew},
  title = {The {VIA} Annotation Software for Images, Audio and Video},
  booktitle = {Proceedings of the 27th ACM International Conference on Multimedia},
  series = {MM '19},
  year = {2019},
%   isbn = {978-1-4503-6889-6/19/10},
  location = {Nice, France},
  numpages = {4},
  doi = {10.1145/3343031.3350535},
  publisher = {ACM},
  address = {New York, NY, USA},
}

@article{ren2015faster,
  title={Faster r-cnn: Towards real-time object detection with region proposal networks},
  author={Ren, Shaoqing and He, Kaiming and Girshick, Ross and Sun, Jian},
  journal={Advances in neural information processing systems},
  volume={28},
  year={2015}
}

@inproceedings{carion2020end,
  title={End-to-end object detection with transformers},
  author={Carion, Nicolas and Massa, Francisco and Synnaeve, Gabriel and Usunier, Nicolas and Kirillov, Alexander and Zagoruyko, Sergey},
  booktitle={European conference on computer vision},
  pages={213--229},
  year={2020},
  organization={Springer}
}

@inproceedings{liu2016ssd,
  title={Ssd: Single shot multibox detector},
  author={Liu, Wei and Anguelov, Dragomir and Erhan, Dumitru and Szegedy, Christian and Reed, Scott and Fu, Cheng-Yang and Berg, Alexander C},
  booktitle={European conference on computer vision},
  pages={21--37},
  year={2016},
  organization={Springer}
}

@article{mmdetection,
  title   = {{MMDetection}: Open MMLab Detection Toolbox and Benchmark},
  author  = {Chen, Kai and Wang, Jiaqi and Pang, Jiangmiao and Cao, Yuhang and
             Xiong, Yu and Li, Xiaoxiao and Sun, Shuyang and Feng, Wansen and
             Liu, Ziwei and Xu, Jiarui and Zhang, Zheng and Cheng, Dazhi and
             Zhu, Chenchen and Cheng, Tianheng and Zhao, Qijie and Li, Buyu and
             Lu, Xin and Zhu, Rui and Wu, Yue and Dai, Jifeng and Wang, Jingdong
             and Shi, Jianping and Ouyang, Wanli and Loy, Chen Change and Lin, Dahua},
  journal= {arXiv preprint arXiv:1906.07155},
  year={2019}
}

@article{dldong2021survey,
  title={A survey on deep learning and its applications},
  author={Dong, Shi and Wang, Ping and Abbas, Khushnood},
  journal={Computer Science Review},
  volume={40},
  pages={100379},
  year={2021},
  publisher={Elsevier}
}

@article{brown2020language,
  title={Language models are few-shot learners},
  author={Brown, Tom and Mann, Benjamin and Ryder, Nick and Subbiah, Melanie and Kaplan, Jared D and Dhariwal, Prafulla and Neelakantan, Arvind and Shyam, Pranav and Sastry, Girish and Askell, Amanda and others},
  journal={Advances in neural information processing systems},
  volume={33},
  pages={1877--1901},
  year={2020}
}

@inproceedings{rombach2022highdiffusion,
  title={High-resolution image synthesis with latent diffusion models},
  author={Rombach, Robin and Blattmann, Andreas and Lorenz, Dominik and Esser, Patrick and Ommer, Bj{\"o}rn},
  booktitle={Proceedings of the IEEE/CVF Conference on Computer Vision and Pattern Recognition},
  pages={10684--10695},
  year={2022}
}

@inproceedings{rematas2022urban,
  title={Urban radiance fields},
  author={Rematas, Konstantinos and Liu, Andrew and Srinivasan, Pratul P and Barron, Jonathan T and Tagliasacchi, Andrea and Funkhouser, Thomas and Ferrari, Vittorio},
  booktitle={Proceedings of the IEEE/CVF Conference on Computer Vision and Pattern Recognition},
  pages={12932--12942},
  year={2022}
}

@inproceedings{he2022masked,
  title={Masked autoencoders are scalable vision learners},
  author={He, Kaiming and Chen, Xinlei and Xie, Saining and Li, Yanghao and Doll{\'a}r, Piotr and Girshick, Ross},
  booktitle={Proceedings of the IEEE/CVF Conference on Computer Vision and Pattern Recognition},
  pages={16000--16009},
  year={2022}
}

@inproceedings{he2016deepresnet,
  title={Deep residual learning for image recognition},
  author={He, Kaiming and Zhang, Xiangyu and Ren, Shaoqing and Sun, Jian},
  booktitle={Proceedings of the IEEE conference on computer vision and pattern recognition},
  pages={770--778},
  year={2016}
}

@article{dosovitskiy2020imagevit,
  title={An image is worth 16x16 words: Transformers for image recognition at scale},
  author={Dosovitskiy, Alexey and Beyer, Lucas and Kolesnikov, Alexander and Weissenborn, Dirk and Zhai, Xiaohua and Unterthiner, Thomas and Dehghani, Mostafa and Minderer, Matthias and Heigold, Georg and Gelly, Sylvain and others},
  journal={arXiv preprint arXiv:2010.11929},
  year={2020}
}

@article{yuan2022volo,
  title={Volo: Vision outlooker for visual recognition},
  author={Yuan, Li and Hou, Qibin and Jiang, Zihang and Feng, Jiashi and Yan, Shuicheng},
  journal={IEEE Transactions on Pattern Analysis and Machine Intelligence},
  year={2022},
  publisher={IEEE}
}

@article{mai2022online,
  title={Online continual learning in image classification: An empirical survey},
  author={Mai, Zheda and Li, Ruiwen and Jeong, Jihwan and Quispe, David and Kim, Hyunwoo and Sanner, Scott},
  journal={Neurocomputing},
  volume={469},
  pages={28--51},
  year={2022},
  publisher={Elsevier}
}

@inproceedings{gaur2018video,
  title={Video annotation tools: A Review},
  author={Gaur, Eshan and Saxena, Vikas and Singh, Sandeep K},
  booktitle={2018 International Conference on Advances in Computing, Communication Control and Networking (ICACCCN)},
  pages={911--914},
  year={2018},
  organization={IEEE}
}

@inproceedings{lin2017feature,
  title={Feature pyramid networks for object detection},
  author={Lin, Tsung-Yi and Doll{\'a}r, Piotr and Girshick, Ross and He, Kaiming and Hariharan, Bharath and Belongie, Serge},
  booktitle={Proceedings of the IEEE conference on computer vision and pattern recognition},
  pages={2117--2125},
  year={2017}
}

@INPROCEEDINGS{icpr,
  author={Adhikari, Bishwo and Huttunen, Heikki},
  booktitle={2020 25th International Conference on Pattern Recognition (ICPR)}, 
  title={Iterative Bounding Box Annotation for Object Detection}, 
  year={2021},
  volume={},
  number={},
  pages={4040-4046},
  doi={10.1109/ICPR48806.2021.9412956}}









\newpage
\appendix
\section{Supplementary material}

\subsection{User experience}\label{ssec:ux}
An IA tool respecting the framework exposed in Figure~\ref{fig:diagram} should be user-friendly. User experience design (UX) studies how to make a user perceive and respond positively to a product. It involves the perception of the utility, the ease of use, and the efficiency of the product. For instance, a door that has a door handle on the ``Push'' side will provide a sub-optimal user experience.
In our \iadet tool, the utility of the tool is made explicit: each time an unlabeled image is loaded, the predictions of the current model are visualized. Thus, as the annotation process goes on, the user can see how the model improves.

For the ease-of-use, we use assisted annotation. The ideal assisted annotation provides a reasonable proposal of bounding boxes that the user can keep and quickly converges to the desired annotation when corrected. In this first version, we only implement the proposal, and corrections of the user are not assisted at all. This works very well when there are few object instances per image. 
Inspired by \href{https://github.com/developer0hye/Yolo_Label}{YOLO-Label}, we depart from the usual click-and-drag bounding box creation, and implement bounding box creation with two clicks: the first for one corner and the second for the opposite corner. These clicks are made with the usual left button. To remove a bounding box only one click is needed. This right-button click will remove the bounding box whose border is closest to it.
All functionalities that are not location specific go through the keyboard: if desired, one can delete all bounding boxes by pressing \verb|Del|. 
The user can change images using keyboard keys that call the \verb|prev| and \verb|next| functions. The annotations are automatically saved when changing an image without the need for user interaction.

Regarding efficiency, model selection is vital. However, we do not expect the user to know about a more efficient model or training method.
Nevertheless, there are a few things that make a tool feel efficient. First, it has to be fast: we solve image loading with \verb|PIL|, while the annotations loading is done with the \verb|pickle| library. Second, the tool has to be responsive: we display the mouse coordinates, which give a feeling of sensitivity when quickly changing as the mouse moves. Third, it has to be informative: the user is shown the tensorboard logs and sees how the loss decreases as the training progresses, and it is also shown the current image path in the context of the other paths in the dataset. One thing that annoys users while annotating is a model that does not respect their annotations. How to fully fit the data is an open question, but we simply make this limitation invisible: the user is always shown the annotations she has made if available.

A few extra elements involve a \verb|Help| button, that displays a short manual and an email address to provide feedback to. The size of the display is also a parameter the user can change. The limitations of the tool involve: (i) the use of \verb|Python tkinter| as User Interface library when \verb|C++| code is presumably faster, (ii) the last image is not saved when closing the tool, (iii) the lack of intra-image assistance, (iv) the absence of zoom-in/out or contrast changing capabilities, (v) the tool is not packaged and (vi) it is launched from the command line, i.e. it is not yet a completely no-code tool.

\subsection{Implementation details}
\begin{itemize}
    \item The chosen network is SSD300 \cite{liu2016ssd}, with the \href{https://github.com/open-mmlab/mmdetection/blob/master/configs/pascal_voc/ssd300_voc0712.py}{configuration file} of PASCAL VOC.  
    \item We modified the training loop to trigger a new dataloader creation at each epoch.
    \item The tool is responsible for using the background model, it makes bounding box predictions for the requested image. In the future, this result could be cached, batched, or pre-run.
    \item The IoU threshold used for the simulation is 0.5
    \item For the simulation, the proposal score threshold was the minimum between $0.7$ and the highest predicted score. In other words, we always predict one bounding box.
\end{itemize}

\end{document}